# BEYOND BIAS AND COMPLIANCE:

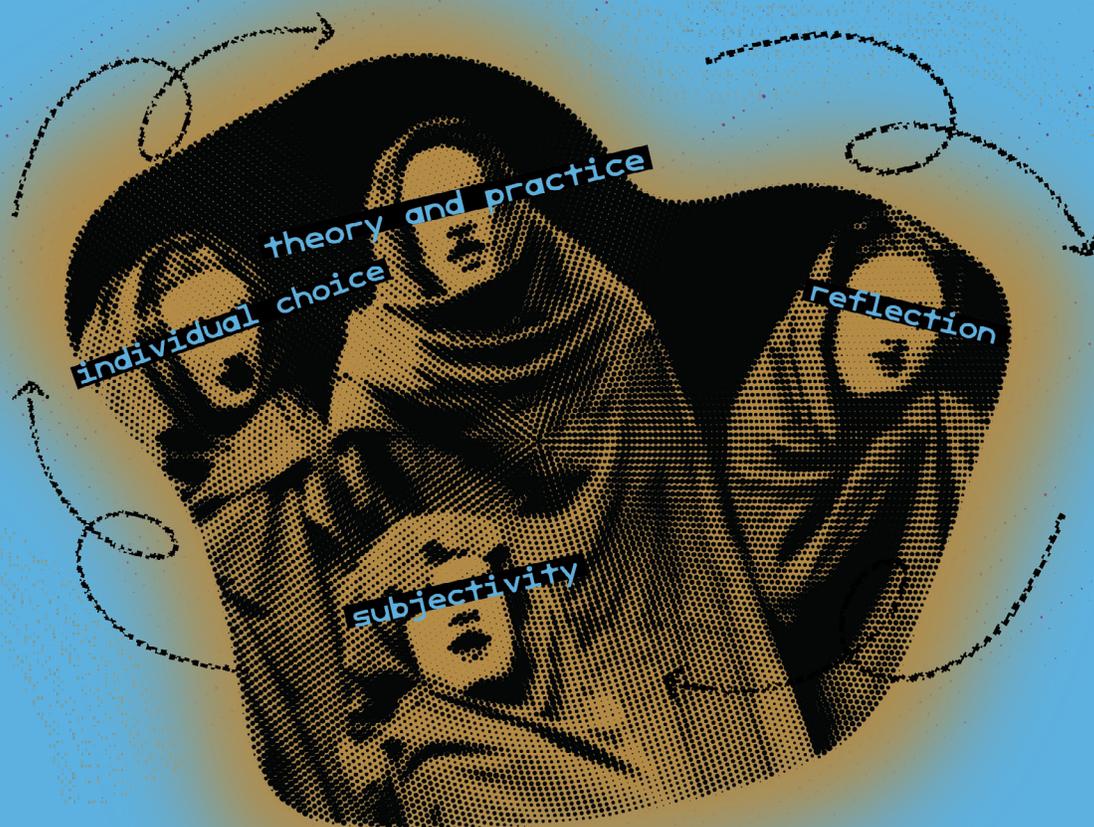

Thomas Krendl Gilbert
Megan Anne Brożek
Andrew Brożek

# TOWARDS INDIVIDUAL AGENCY AND PLURALITY OF ETHICS IN AI

daios

# TABLE OF CONTENTS

# EXECUTIVE SUMMARY

AI ethics is an emerging field with multiple, competing narratives about how to best "solve the problem" of building human values into machines. Two major approaches are focused on bias and compliance, respectively. But neither of these ideas fully encompasses ethics: using moral principles to decide how to act in a particular situation.

In this document, we first evaluate multiple AI ethics companies, many of which have started from bias, compliance, or another vague understanding of ethical principles. None of these solutions are value-agnostic. We next present a reflective interlude on the limitations of current AI ethics approaches, leading to a new, alternative method.

If removing bias is a false start, then what? The daios method posits that the way data is labeled plays an essential role in the way AI behaves, and therefore in the ethics of machines themselves. The argument combines a fundamental insight from ethics (i.e. that ethics is about values) with our practical experience building and scaling machine learning systems (i.e. data determines model behavior). We want to build AI that is actually ethical by first asking foundational questions: how can we build good systems? What does "the good" mean? And who should determine it?

We also provide a deep dive into the philosophical foundations of daios. The product itself is not set in stone and will likely change after the publication of this whitepaper. However, our philosophical starting points will not. These foundational first principles stem from multiple traditions of philosophy, particularly from the works of Aristotle, Søren Kierkegaard, Hubert Dreyfus, and Friedrich Nietzsche.

Our values are the theoretical starting points for the design of our product. Currently, daios is a platform that teaches machines morality by co-creating algorithms with end users. It builds on top of platforms that already use AI, connecting end users and development teams through data, as a value-neutral adjudicator.

We also include two appendices: the first contains our glossary of key technical terms, and the second includes our interviews with development teams working to improve performance benchmarks for deployed AI systems.

Building ethical AI creates a foundation of trust between a company and the users of that platform. But this trust is unjustified unless users experience the direct value of ethical AI. Until users have real control over how algorithms behave, something is missing in current AI solutions. This causes massive distrust in AI, and apathy towards AI ethics solutions. The scope of this paper is to propose an alternative path that allows for the plurality of values and the freedom of individual expression. Both are essential for realizing true moral character.

While AI ethics remains an underserved market, we are building a company that captures that untapped market share. As such, this white paper has been written for CEOs and CTOs of AI/ML startups who are seeking an alternative to the current regime of AI ethics and want to try something new. We aim to reach investors and venture capitalists who focus on AI/ML and wish to learn more about our methods.

# THE PRESENT LANDSCAPE OF AI ETHICS IN INDUSTRY

The number of companies pursuing ethical AI has grown rapidly. While academic research into AI ethics is ongoing[1], here we track the promise and limitations of services on the market now.

Table 1 presents this landscape. The columns capture the stated goals of the company in terms of key outputs and deliverables:

PRACTICAL: concrete outputs or material tools that can be integrated into machine learning pipelines.
THEORETICAL: concepts, lexicons, or research papers intended to improve fundamental understanding of AI ethics problems.
BOTH: a mix of applied and abstract deliverables that together fit the company's own qualitative views of what is missing in AI ethics.

The rows capture the methods a given company pursues to enact those ends:
ETHICAL PRINCIPLES: Codification of institutional roles, standards, or principles by which whatever is being built would be recognizably ethical and responsible to stakeholders.
NEW TECHNOLOGY: Any designed tools (interfaces, optimization techniques, simulations, etc.) that are interpreted as necessary pieces of infrastructure for the wider project of developing ethical systems.

|  | PRACTICAL | THEORETICAL | BOTH |
|---|---|---|---|
| Ethical Principles | Frameworks and governance (e.g. Credo AI, Parity) | Ethics research initiatives and non-profits (e.g. All Tech Is Human) | Civil society advocates (e.g. DAIR, Data & Society) |
| New Technology | AI development tools (e.g. Arthur, Holistic AI, Fairplay) | AI Alignment (e.g. OpenAI, Anthropic, Aligned AI, Conjecture, Preamble) | Open source tools (e.g. Hugging Face, Cohere) |

Table 1: Distinct means (rows) and ends (columns) in companies' pursuit of ethical AI.

Below we present each of these cells in greater detail, including the assumptions among representative companies about how AI could be made ethical.

### FRAMEWORKS AND GOVERNANCE
Many companies focus on making sure that particular development teams are compliant with external standards or rules, such as Responsible AI[2] or Explainable AI[3]. For example, Credo AI's core product is providing software-as-a-service to support AI governance risk assessment at scale[4]. A team's use of that service is interpreted as compliance with these rules, and thereby treated as a good proxy for desirable system behavior and model outputs. Another example is Parity, which provides a platform for companies to adhere to present and upcoming AI regulations like the EU AI Act[5].

---

[1] Representative findings and publications are available in the proceedings of the ACM conference on Fairness, Accountability, and Transparency (ACM FAccT) and the AAAI/ACM conference on Artificial Intelligence, Ethics, and Society (AIES).
[2] See for example here: https://www.pwc.com/gx/en/issues/data-and-analytics/artificial-intelligence/what-is-responsible-ai/responsible-ai-practical-guide.pdf
[3] See for example the IBM approach here: https://www.ibm.com/watson/explainable-ai
[4] https://www.credo.ai/lens
[5] https://www.getparity.ai/

However, AI systems with the most advanced AI capabilities now work at massive, often unprecedented scales. This means changes related to compliance must also be implemented at scale. Frameworks by their very nature are laborious and slow, making them much better suited to narrowly-scoped AI applications like credit scoring or facial recognition whose capabilities are more fully understood by designers prior to deployment. Moreover, this "checklist approach" to governance becomes less and less effective when engineers want to respond and change AI behavior quickly. This is because checklists are static[6] –they must be applied to a self-contained data distribution or set of model outputs. Yet many AI harms arise from the dynamics generated by an AI system as it continuously operates in the real world. For example, engineers regularly retrain the recommender algorithms used on social media. This makes the feedback loops between human data and recommended content much more pressing and significant than any particular data structure. To remain effective at scale, governance frameworks for compliance must be integrated within these feedback loops, not sit outside them.

### AI DEVELOPMENT TOOLS

Another approach is to create new tools that shed light on design considerations or specific system components such as models or data. For example, Arthur[7] strives to be a proactive model monitoring platform that tracks whether or how AI deployments are performing as expected. Holistic AI[8] provides assessments on individual system components "at any level of maturity or scale" in order to minimize liability concerns. Fairplay[9] provides "fairness as a service" for financial institutions, using AI to inspect a given company's own automated decisioning models for potential forms of bias. Other companies are assisting in traditional ML operations. For example, Weights & Biases[10] works to speed up development time through the use of experiment tracking, dataset versioning, and better model management so that developers have a better sense of what is going on with their own pipelines in real time.

These companies assume that measuring the behavior of AI models is a path to more ethical AI. Of course, this approach is only as effective as the measurement techniques being used and the thresholds for harm or risk that have been given–if either is absent, lacking, or inappropriate, these tools are ethically useless. Moreover, even if the measurement is accurate and thresholds are well-defined, development tools of this kind cannot directly alter AI behavior. Say, for example, a diagnostic tool reveals that an algorithm recommends inappropriate content to social media users. On its own, this information is insufficient to fix the algorithm, because it is not clear how to tie this emergent behavior back to specific design choices. Is the problem that content creators have learned to "game" the algorithm? Can the system for some reason not recognize that certain types of content are inappropriate? Are human moderators not able to do their job properly? Flagging the problem is not enough; a substantively ethical approach must suggest alternative design interventions based on why the problem exists.

### ETHICS RESEARCH INITIATIVES AND NONPROFITS

These organizations develop the concepts, methods, frameworks, initiatives, and assumptions about AI ethics needed to develop safe and beneficial AI systems. As such these organizations usually generate research as their core output, rather than commercial products. A good example is All Tech Is Human, which works to build a "pipeline" of participants and research agendas that are "diverse, multidisciplinary, and aligned with the public interest".[11] Here the goals are that AI ethics are made achievable by "changing the people involved in it" rather than focusing on the development of core system components. daios does not consider these types of organizations to be direct competitors, since they are not creating practical components (or view their creation as secondary to basic forms of community-building).

---

[6] Checklists may be updated sporadically or on an annual basis, but these updates are decoupled from the behavior of the systems they are intended to regulate.
[7] https://arthur.ai/
[8] https://holisticai.com/
[9] https://fairplay.ai/
[10] https://wandb.ai/site
[11] https://alltechishuman.org/about

### AI ALIGNMENT

These companies try to achieve fundamental technical breakthroughs for the creation of "aligned" AI, i.e. systems that share or reflect the structure of human values. As part of this mission, they are committed to building artificial general intelligence (AGI) that is "safe" or "friendly" rather than merely ethical–the system's behavior will avoid "existential risks" and provide demonstrable benefits to humans.

While both the concept and technical development of AGI remain inchoate, companies in this space (for example OpenAI[12], Anthropic[13], Aligned AI[14], Conjecture[15], and Preamble[16]) have already deployed technologies that act as prototypes of more general capabilities. These include large generative models like GPT-3 and DALL-E 2, both of which were developed by OpenAI. A major outstanding question for these companies is how to develop systems of unprecedented scale and complexity, often requiring millions of dollars of outside investment and external funding, in ways that are verifiably ethical or accountable to stakeholders.[17] This requires the development of toolkits or APIs for those systems that are both usable and robust to misuse by nefarious or untrustworthy actors who could contort the behavior or use of these systems to be unsafe.

### CIVIL SOCIETY ADVOCATES

Some organizations, like Distributed AI Research (DAIR[18]) and Data & Society[19], concentrate on identifying and diagnosing the types of harms that AI systems commit on human populations. The data-driven optimization at the heart of today's AI systems often causes unintended effects on society over time, and these institutes work to expose these processes and hold them accountable. This mission is both theoretical and practical, as effects must be both investigated and mitigated by raising public awareness to hold systems responsible for their effects.

This advocacy is not strictly critical of AI. DAIR researchers are now building AI tools and interfaces that are intended to replace the most pernicious components of industry-scale systems, and use AI as a forensic tool to track societal harms in new ways.[20] These include language technology that serves marginalized communities, as well as low resource settings for AI.[21] Meanwhile, Data & Society has proposed new tools for improving how the properties and risks of AI systems are documented for public benefit. Beyond critique, these organizations are working to reform the industry practices behind how AI is now built and deployed in human contexts.

### OPEN-SOURCE TOOLS

Companies such as Hugging Face[22] and Cohere[23] have drawn attention to building more ethical AI models by making them either more accessible to developers or fully open source, rather than strictly more capable or provably safe (as in AI alignment). The core assumption is that making systems open-source will encourage involvement in their development, making them better-aligned with community norms. However, this assumption is restricted to model training and outputs. In particular, there is not yet a corresponding open-source commitment to data, which–as recently emphasized by Andrew Ng[24] –has the single largest impact on model behavior. Data creates ground truth for machine learning systems, establishing for the machine what is real and not real. Models then conform to trends in data. Beyond efforts at data documentation, the labeling process behind data remains largely unexplored as a key vector for making AI more ethical.[25]

---

[12] https://openai.com/
[13] https://www.anthropic.com/
[14] https://www.aligned-ai.com/
[15] https://www.conjecture.dev/
[16] https://www.preamble.com/
[17] See for example Brundage, Miles, et al. "Toward trustworthy AI development: mechanisms for supporting verifiable claims." *arXiv preprint arXiv:2004.07213* (2020).
[18] https://www.dair-institute.org/
[19] https://datasociety.net/
[20] https://www.dair-institute.org/research
[21] https://openreview.net/forum?id=WV0waZz9dTF
[22] https://huggingface.co/
[23] https://cohere.ai/
[24] https://fortune.com/2022/06/21/andrew-ng-data-centric-ai/
[25] There are also many competitors within the data labeling sphere, such as Scale AI and Amazon SageMaker Data Labeling, among others. As of December 2022, none of these strictly data labeling services are focused on how ethics of algorithms relates to data labeling for machine learning.

# INTERLUDE: THE NEED FOR AN ALTERNATIVE

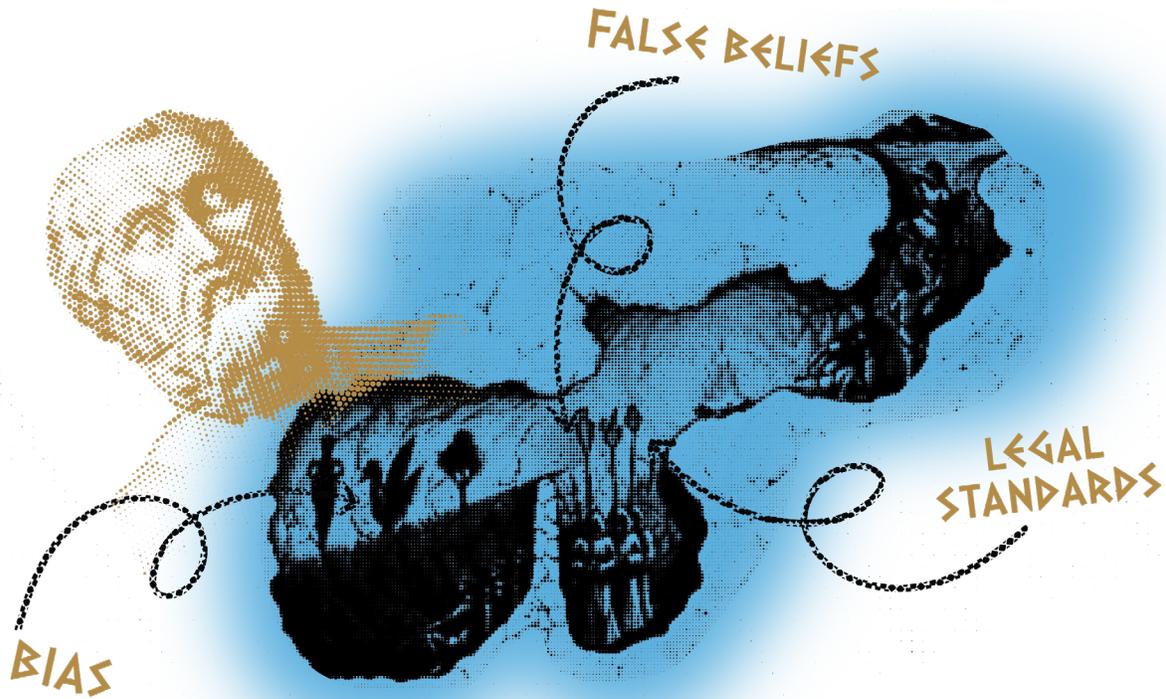

The present configuration of the AI ethics landscape. Like Plato's cave, companies deploy biased representations of reality derived from legal standards, instead of asking what is truly ethical

In late 2021, while searching for a co-founder, one of us met an ivy league graduate with a background in philosophy and computer science. Sounds perfect, right? During our discussion, she mentioned something revealing: "If you want to start with data, you need to gather data from all perspectives, from every possible perspective in the world. That's how you would solve the problem of AI ethics."

This interaction reflected an important limitation of present AI ethics "solutions": they do not consist of much actual ethics. Many AI ethics practitioners approach the problem of aligning machines with ethical values from a psychological, sociological, or legal perspective. Although these domains bring invaluable insight, psychology, sociology, and law are not ethics. The focus on bias and compliance as the main thrust of most ethical AI is a symptom of this problem.

### 1. ETHICS AS COMPLIANCE
Ethics is a branch of philosophy that determines right and wrong by analyzing and defending different concepts of values. But AI ethics as understood by many AI ethics practitioners, startups, and corporations, lacks a proper understanding of ethical values. The current market frames ethics as a legal issue (e.g. compliance with the law) or uses values that are arbitrary and vague (e.g. buzzwords such as responsible, safe, etc.). Compliance with present legal standards and prospective government regulations is necessary, but not sufficient, as this perspective pigeonholes ethics into *what is permissible*, rather than *what is good*.

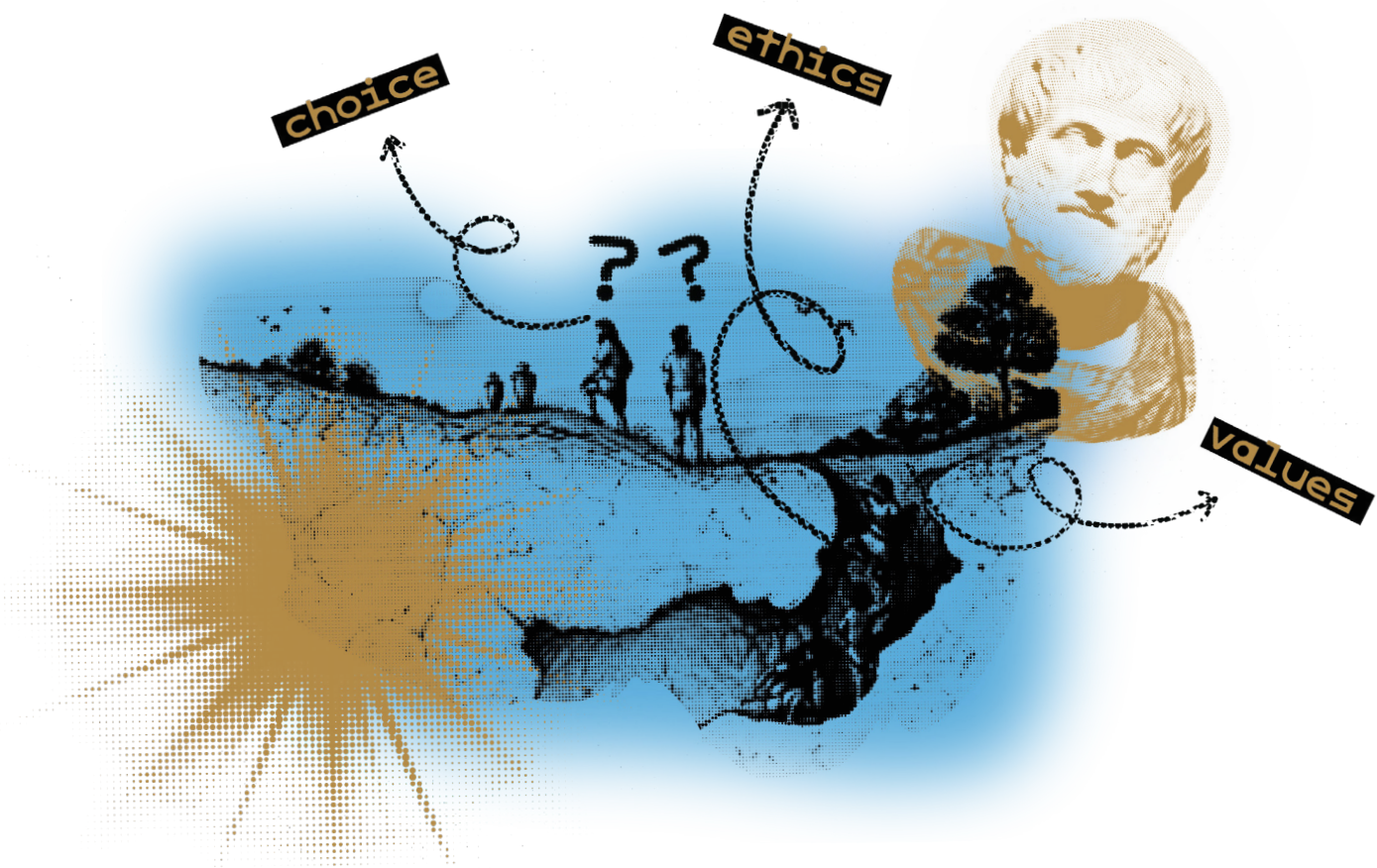

Our perspective, influenced heavily by Aristotle, emphasizes the need to move beyond what is legal and inquire into what it is actually ethical for AI systems to do or not do.

The rigidity of ethics-as-compliance is problematic for four reasons. While necessary, compliance defines the problem space of AI ethics in ways that are:

1. *inflexible*, excluding new ethical problems as they arise;
2. *narrow*, defining 'ethics' as conformity with standards determined by governing bodies;
3. *impractical*, as the algorithm will inevitably make mistakes when applied to new data;
4. *harmful*, as it prevents the emergence of dissenting ethical values over time;

On the contrary, ethics is an issue that belongs in civil society, and requires deliberating about what is right and wrong. Conflating vices with crimes and virtue with conformity leads to unhappiness and civil strife.[26] Once the notion of values as predetermined is abandoned, the door opens for a more proactive approach that remains values-agnostic.

### 2. ETHICS AS MERELY TECHNICAL
Another popular approach to AI ethics has been for engineers to develop tools for more ethical AI/ML operations. This frames ethics as a technical problem to be solved, instead of a question to be posed and answered via reflection.

The best engineering best practices rightly strive to ensure standardized, repeatable, and measurable algorithms. But ethical AI/MLOps companies often build based on generally accepted definitions of a "good ethical system", which are typically vague and unclear. Solutions in this category include values such as "fairness", "responsible", and "bias". The question becomes "Fair to whom?" "Responsible to whom?" "Biased against whom?" The answer will differ according to different individuals, in different contexts. Ultimately, the problem of "whose ethics should be built into this system" is pushed down the road. Solving "Good to whom?" takes philosophical foresight and appropriate reflection. It cannot be resolved in a quick google search.

---

[26] See for example Spooner, Lysander. *Vices are Not Crimes: A Vindication of Moral Liberty*. Classics Press, 2010.

### 3. ETHICS AS MERELY THEORETICAL

Finally, some companies approach AI ethics in a research-only manner, combining philosophy, computer science, economics, and ethics. These companies may be concerned with long-term AI ethics, such as AI as an existential threat.

However, AI/ML systems are already being deployed by companies all over the globe, in unprecedented contexts. A solution is required for systems that are being deployed right now, not ones that may exist in 50 years. The problem with separating high-level theory and engineering practice is that many research questions become speculative and unanswerable in the current paradigm. Researchers slowly lose the intuition gained by touching a thing, experiencing a thing, or smelling a thing. Very few AI ethics startups successfully combine research and practice.

### 4. AN ALTERNATIVE FRAMING OF THE MARKET

The AI ethics market is a combination of governance, MLOps, and research companies. There are a few AI ethics offerings that define ethics beyond these framings[27], such as companies like Hugging Face and Holistic AI. However, these AI ethics startups are focused on models rather than training data. Data creates ground truth for machine learning systems, establishing to the machine what is real and not real. Models conform to trends in data. We must ask ourselves: why are we playing with model parameters rather than examining the power of training datasets?

With a proper understanding of ethics, the market widens beyond building AI to avoid certain biased results. There is also a need to teach AI certain values, and take a more *active* stance in algorithm creation. This active relationship with AI is the vision of daios.

The ultimate question becomes: can AI actually be taught what is good? We believe it is possible once two basic concepts are combined from ethics and machine learning: labeled data and ethical values.

[27] Data-focused AI ethics solutions also exist in the form of protecting a brand, or brand safety, which requires sifting through explicit training data to censor AI from producing certain outputs. The policies created by companies may be decided by a federation (e.g. GARM as within advertising).

# THE DAIOS APPROACH

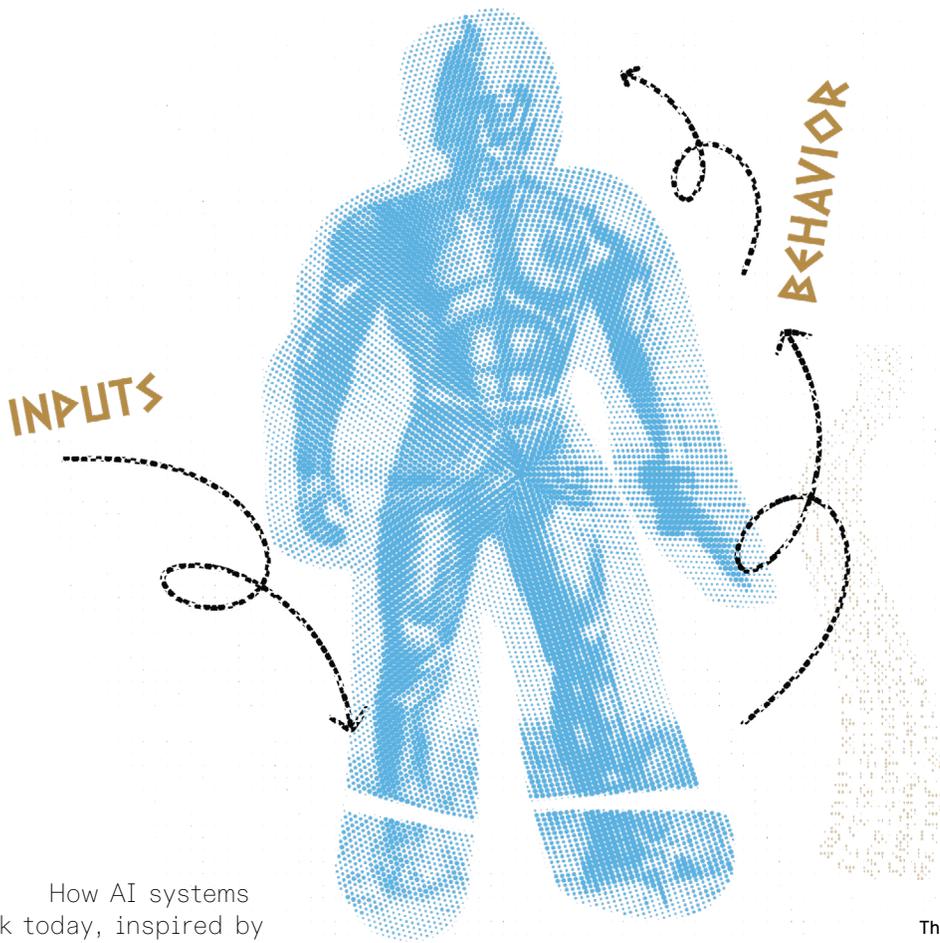

This image adapted from the film Jason and the Argonauts (1963)

How AI systems work today, inspired by the myth of Talos from Greek mythology. AI systems take in data inputs and then take actions or behave according to how the data is structured.

daios method departs from the existing approaches following from three fundamental truths:
O     Data determines truth for machines.
O     Ethics is not about compliance or removing bias, but acting based on what is good.
O     Building ethical AI/ML systems first requires a neutral adjudicator to identify ethical values within those systems.

Now, we will go deeper into the first point.

### 1. WHAT IS MACHINE LEARNING?
Machine learning is a method of computer science that uses data to create a probabilistic representation of the world in the form of a mathematical function or algorithm. Data are pieces of information (i.e. images, text, sound bytes, numbers) upon which computers are able to operate. As Thomas Mitchell (1997) famously put it:

*"A computer program is said to **learn** from experience E with respect to some class of tasks T and performance measure P, if its performance at tasks in T, as measured by P, improves with experience E."*

When we say that a machine learning model learns, we mean that the machine learning model is inferring patterns from the data. A computer is able to learn a particular task by training it on any number of finite (thousands, millions, quadrillions) examples.

## 2. WHAT IS DATA FOR MACHINES?

Data provides ground truth for machines, the first fundamental truth as stated at the beginning of the section.

Ian Goodfellow, Yoshua Bengio, and Aaron Courville (2016) speak of transitioning away from the *knowledge base*, an approach to AI that involved the hard-coding of knowledge about the world into formal languages. This process proved to be unwieldy. People struggle with creating formal rules with enough complexity to accurately describe the world. So, AI needed a way to acquire its own understanding of the world.

> The difficulties faced by systems relying on hard-coded knowledge suggest that AI systems need the ability to acquire their own knowledge, by extracting patterns from raw data. This capability is known as machine learning. The introduction of machine learning enabled computers to tackle problems involving knowledge of the real world and make decisions appear subjective.[28]

Machines read raw data and understand that information to be reality. Machines understand data literally, not metaphorically. Currently, issues with machine learning models are often resolved by adding more data to training datasets or removing problematic data.

## 3. WHAT IS THE BEHAVIOR OF MACHINES?

The output of machine learning models over time determines the behavior of machines. We can analyze machine behavior as actions, which will be important to the definition of ethics in section 5.

## 4. WHAT IS PHILOSOPHY?

Philosophy is the pursuit of truth. The purpose of philosophy is to ask, "why?" For example, "why do we do science? Should we have limitations on research — if yes or no, why?"

Both science and philosophy aim at truth. However, science is concerned with knowledge that can be verified (or falsified) through the senses. Although philosophy and science are not necessarily opposed, philosophy often goes beyond the questions that science is able to ask or answer.

## 5. WHAT IS ETHICS?

Ethics is a branch of philosophy that tries to determine right and wrong by analyzing and defending concepts of value. Ethics also involves the analysis of actions as allowed or not allowed by particular value systems. This is the second fundamental truth of the daios method.

Ethics is often confused with issues of legality, social convention, politics, and religion. In reality, however, ethics is truly about discovering the right thing based on reason and context. Following the law may make you a lawful person, but not necessarily an ethical one. Laws are created by humans, who may be flawed in their judgment, or may exist due to an archaic idea that no longer applies to the current context. Politics allows citizens or rulers to determine a given course of action for society, but not to articulate reasons why that course is right or wrong. Social convention is dependent on habit and religion is dependent on divine revelation, rather than human reasoning.

## 6. WHAT DOES THIS HAVE TO DO WITH MACHINE LEARNING?

A machine executes based on what it has been told to do, dictated in the form of programming. In the case of machine learning, a machine is programmed to learn based on example tasks in the form of training data. As previously stated, machines understand data to be a literal representation of the world.

---

[28] Goodfellow, Ian, Yoshua Bengio, and Aaron Courville. *Deep learning.* MIT press, 2016.

However, machines do not analyze their own actions as right or wrong according to a value system. Since acts performed by a machine are also allowed by the machine, machine behavior *implicitly assumes all actions committed by the system are "ethical"*. While humans are able to distinguish between actions that are descriptive (you observe something) or normative (you say it shouldn't happen), machines do not naturally do this.

Some AI ethicists argue that collecting more data will produce better actions since more data means a more complete view of the world. For example, the theory of metanormativism holds that moral dilemmas will disappear once enough data is collected to resolve ambiguities between equally appealing or unappealing options.[29] However, this approach assumes that machines will always act in ethical ways, which, as we know from the limitations and failures of large language models, is not the case.[30]

## 7. WHAT ARE THE NEXT STEPS?

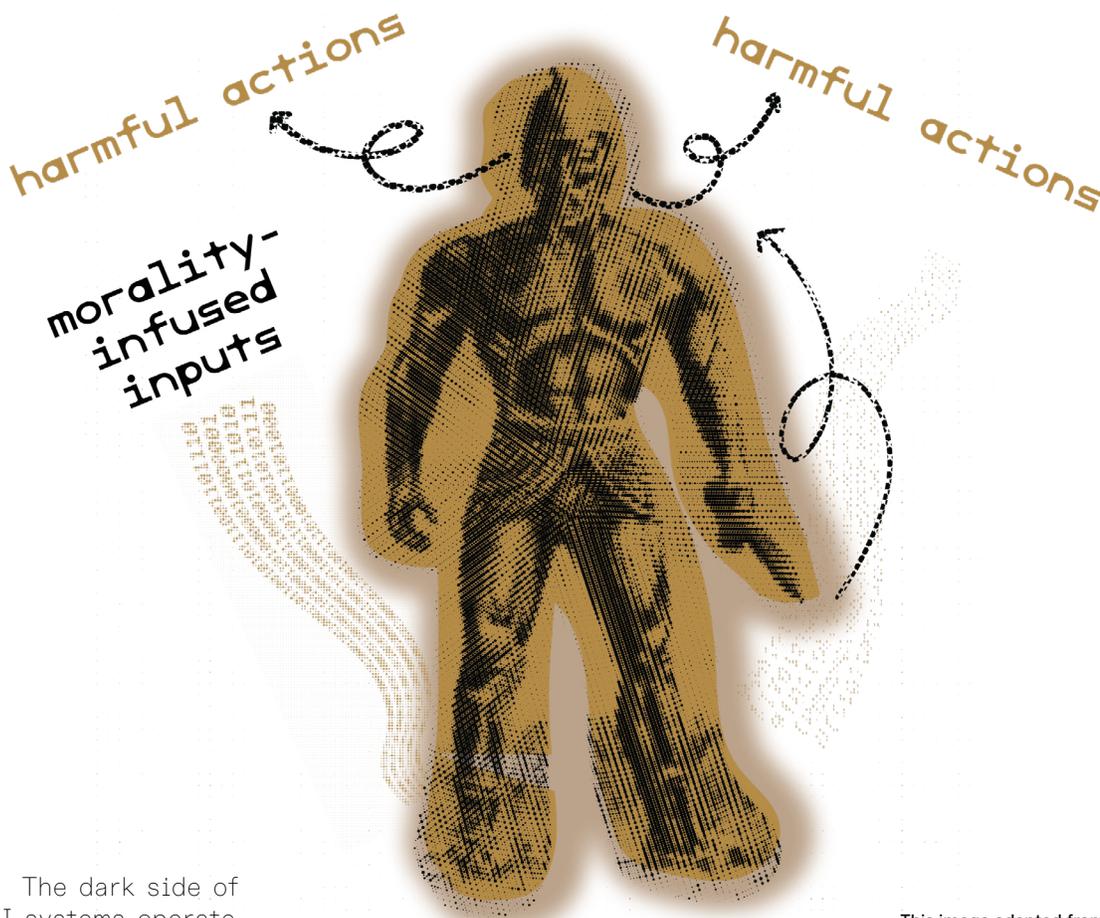

This image adapted from the film Jason and the Argonauts (1963)

The dark side of how AI systems operate. In the current paradigm, morality-infused data is fed into the system, which then learns to take actions that are harmful or violate ethical criteria.

Since machines are misbehaving in unethical ways, we must consider solutions that are able to alter the behavior of current AI systems. We already know that data determines the behavior of machines. If we wish to make machines ethical, we must start with data. We also must establish a method for determining what is 'good' or ethical, since we cannot rely strictly on legal precedent or social convention. It is possible to add another layer within data, an *ethical layer*.

[29] MacAskill, William. Normative uncertainty. Diss. University of Oxford, 2014.
[30] Bommasani, Rishi, et al. "On the opportunities and risks of foundation models." *arXiv preprint arXiv:2108.07258* (2021).

# AXIOMS FOR ETHICAL AI

> And what we said above will apply here as well: what is proper to each thing is by nature best and pleasantest for it; for a human being, therefore, the life in accordance with intellect is best and pleasantest, since this, more than anything else, constitutes humanity. So this life will also be the happiest.
> — Aristotle, Book X, Chapter 7, *The Nicomachean Ethics* [31]

> To know what actions are virtuous, and what vicious — in other words, to know what actions tend, on the whole, to happiness, and what to unhappiness — in the case of each and every man, in each and all the conditions in which they may severally be placed, is the profoundest and most complex study to which the greatest human mind ever has been, or ever can be, directed.
> — Lysander Spooner, *Vices are Not Crimes: A Vindication of Moral Liberty* [32]

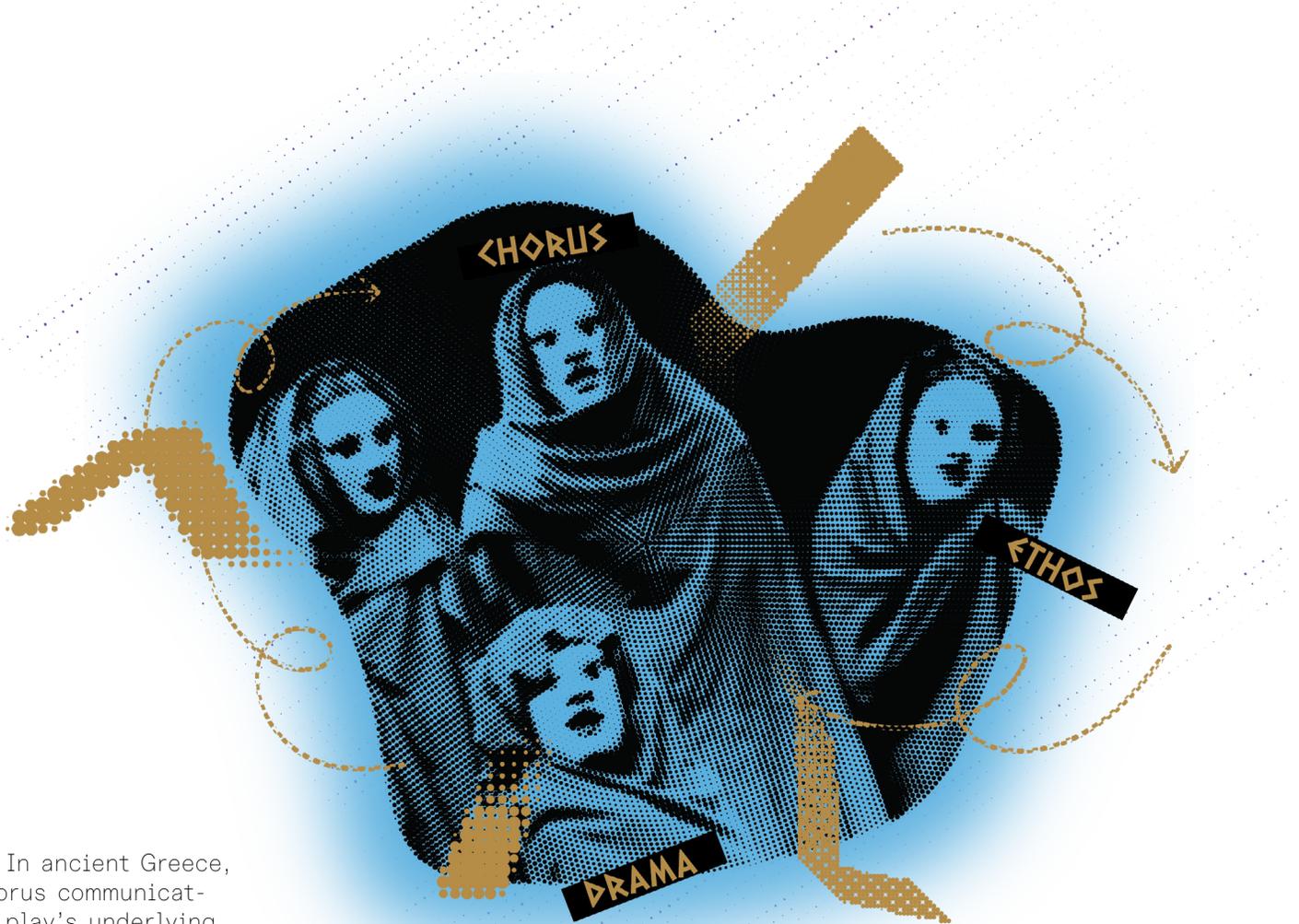

In ancient Greece, the chorus communicated the play's underlying ethos, balancing the dramatic forces of the play against the emotions felt by the audience.

[31] Crisp, Roger, ed. *Aristotle: Nicomachean Ethics*. Cambridge University Press, 2014.
[32] Spooner, Lysander. *Vices are Not Crimes: A Vindication of Moral Liberty*. Classics Press, 2010.

13    AXIOMS FOR ETHICAL AI

Our intuitions about right and wrong, our knowledge of the world itself, all arise from specific contexts. They may be inherited from one's culture. They may be personal, in which case only individual people have access to them. In either case, criteria for virtuous behavior come from people–they are not otherworldly and do not come as a universally known fact. These contexts are largely absent from the data powering the systems built by present AI ethics companies.

How can contexts be accounted for? In this section we highlight the four key axioms that guide the daios approach to excavating moral contexts and integrating them with AI systems. Some of these axioms are quite simple, even obvious. But unless AI's builders can name and confront them, AI systems will remain mere shadowgraphs of human values, reflecting only their form and little of their content.

### 1. TECHNOLOGY OBSCURES THE MORAL CHARACTER OF INDIVIDUALS.

One fundamental assumption at the foundation of daios is that technology obscures the individual, and therefore, moral choice.

Martin Heidegger was the first to critique automated machine technology in his essay in 1954 on *The Question Concerning Technology*, after observing the use of self-guided missiles in WWII. Mark C. Taylor emphasizes that "Heidegger saw the ways in which these technologies were taking the human individual and the decisiveness of the individual out of the loop… Making a human being a function of larger systems." Automated machine technologies were being used in a way that completely neglected individual input, abstracting away the need for deliberation and choice by those involved with the automated process. Heidegger adapted his critique of automation excluding the individual from Søren Kierkegaard, who wrote in response to Hegel.

Hegel argued for a worldview that evaluated history in terms of a *dialectic*, which would oscillate between a thesis and an anti-thesis, pushing world development along to a specific endpoint called the end of history. In this view of reality, the particular human individual does not have significance. The world develops at a completely abstract level, leaving the identities of specific people, or any other details, as indeterminate. For Hegel, humans are a function of the system; in many ways, AI technology is Hegelian, with individuals losing their ability to participate in the system as specific entities but only as sources of data in the totality of the whole AI feedback loop. Individuals cannot express their identity, agency, and choice.

This is in opposition to Kierkegaard, who believed that the individual is primary and the system is secondary. Our identity is determined when we stand alone, in isolation from other individuals. One may imagine Abraham, alone, on Mount Sinai, deciding whether he should sacrifice his first-born son or betray God. Kierkegaard was the first to realize the impact of modern mass media, as he himself was greatly affected by the creation of the printing press.

Today's machine learning algorithms are so complex and fast at making decisions that humans are unable to intervene at the moment of decision. For example, many automated vehicle (AV) companies involve human operators to deal with a human able to take manual control in an emergency. Switching control to a human in risky situations solves the problem, but it becomes inefficient for humans to participate at scale. The human operator also lacks the context of the decision, which is especially problematic when additional milliseconds could change the outcome of the situation.

Taylor says that humans have already created a *cybernetic relationship* with all AI technologies, such as when we interact with AI on our phones, on websites, or any other technology. The relationship between humans and AI is able to become a positive feedback loop, where we are creating the technologies that (re)create us.

Unfortunately, these positive feedback loops do not yet exist in AI.[33] As Heidegger writes, "A craftsman is not aware of a tool until it breaks." We only realize that something is awry after AI misbehaves. When mistakes are made that cannot be reversed and at a speed that is hard to keep up with, you realize that you are, in fact, not part of that system in a meaningful way. Your individual identity is lost within an abstract system. You do not have agency when you use AI. No one does.

### 2. PRACTICE AND THEORY SHOULD BE INTERTWINED.

Another first principle behind our methodology, core business decisions, and, ultimately, day-to-day operating is that theory and practice need each other in order to work.

Working as a unified team is central to innovation, but the question is how teamwork can produce the creative and thorough results necessary to build ethical AI. Building AI to be aligned with specified ethical values combines multiple domains of knowledge. Beyond specialized fields of inquiry (computer science, engineering, philosophy, ethics, economics), this also includes emotional intelligence, practical skills, technical proficiency, and personal life experience.

Theory and practice can be combined in many ways. One concrete example is the historical development of the transistor. In order to replace the vacuum tube, Bell Labs brought together a team of skilled researchers including William Shockley and John Bardeen, both experts in quantum theory, and Walter Brattain, a deft experimentalist. Shockley and Bardeen would theoretically reason through problems, then, Brattain brought their ideas to life through experimentation.

Quick iteration gave immediate feedback, allowing researchers to brute force walk through multiple scenarios and invalidate hypotheses efficiently. The problems the team was trying to solve were hard and, naturally, most experiments did not succeed. However, the process gave vital information for potential paths forward.

The cycle of theory building, experimentation, and theory revising is essential for any empirical science. Although computer science "focuses on the theory and technology of computation itself"[34], every algorithm in computer science needs to be run on something. Software engineering is the empirical correlate that brings computer science to life, through the cyclical relationship of theory and practice.

Physical proximity was also crucial to the creative environment within the transistor team. Bardeen and Brattain had the habit of sitting side by side when inventing the transistor, with Bardeen offering ideas and Brattain trying them out.

However, theory did not always determine practice: "Usually Bardeen's theories led to Brattain's experiments, but sometimes the process worked in reverse: unexpected results drove new theories… they acted out the old physicist joke: they knew that the approach worked in practice, but could they make it work in theory?" (143). While reality did not always fit nicely into Bardeen's theories, innovation occurred in the spontaneity of the moment, and the results were immensely powerful.

Indeed, the philosopher of science Paul Feyerabend argues that theory and practice are often two sides of the same process and should occur simultaneously: "Creation of a thing, and creation plus full understanding of a correct idea of a thing, are very often parts of one and the same indivisible process, and cannot be separated without bringing the process to a stop. The process itself is not guided by a well-defined programme, and cannot be guided by such a programme, for it contains the conditions for the realization of all possible programmes" (10). Not only are theory and practice meant to be combined, but if they are separated then the whole process will be thwarted.

---

[33] Dobbe, Roel, Thomas Krendl Gilbert, and Yonatan Mintz. "Hard choices in artificial intelligence." *Artificial Intelligence* 300 (2021): 103555.

[34] https://www.catalog.caltech.edu/documents/128/catalog_22_23.pdf

It's tempting to credit the process to rational interchange between theory and practice, but in reality, the urge to innovate is prior to or beyond the rational:

> [The process] is guided rather by a vague urge, by a 'passion' (Kierkegaard). The passion gives rise to specific behavior which in turn creates the circumstances and the ideas necessary for analyzing and explaining the process, for making it 'rational'… Theories become clear and 'reasonable' only *after* incoherent parts of them have been used for a long time. Such unreasonable, nonsensical, unmethodical foreplay thus turns out to be an unavoidable precondition of clarity and of empirical success (10–11).

The creative process must be given space to realize its vague urges, to play with theories that may not make sense and chase insight, in order to eventually come to truths that can be made sense of after the fact.

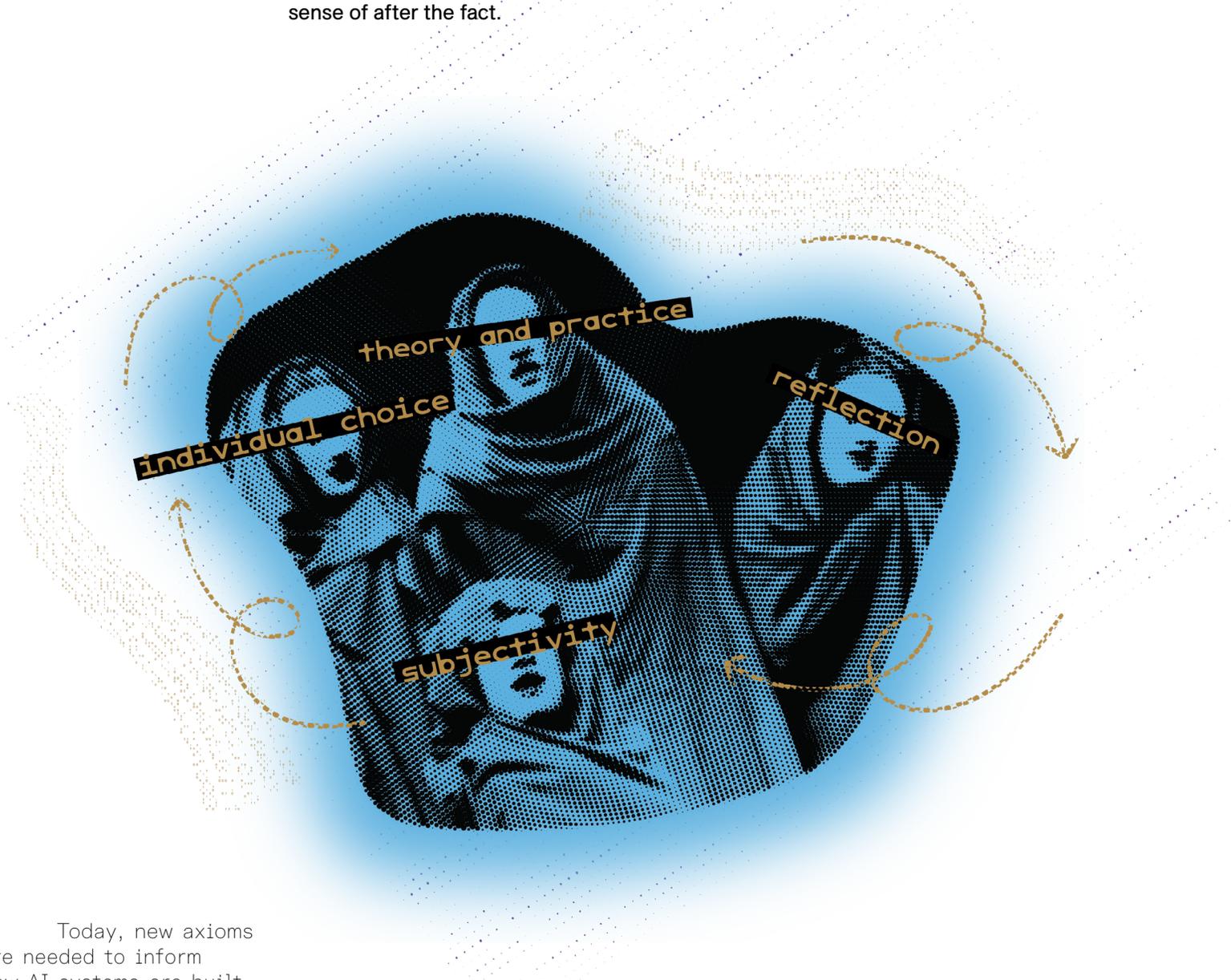

Today, new axioms are needed to inform how AI systems are built. While sometimes in tension with each other, these axioms help comprise a new ethos for AI development.

### 3. BEING ETHICAL REQUIRES ACTING WITH INTENT.

Decades ago, Hubert Dreyfus argued that AI systems could never be as capable as humans. This is because human activity is oriented to the world–we touch things, and grapple with them in pursuit of goals that are meaningful to us. Humans have things they care about and want to defend. AI, however, is oriented to symbols–rules, lines of code, and representations that are assumed to map onto reality. For Dreyfus, AI is doomed to fail because there will always be slippage between the world and the symbols used to refer to it.[35]

Today's AI systems now frequently meet or surpass human capabilities, from driving cars to generating images or text, and no longer rely on explicitly-coded rules. Still, AI may find itself in situations where it is not clear what the right or best thing to do is. Consider the case of a spouse rushing to get their wife to the hospital after she goes into labor. They might decide to speed through red lights, or call an ambulance, or instead prepare to deliver the baby themselves at home. The point is that they must evaluate possible actions in light of what is most important–the safety of their wife, the health of the infant, or following the law. Without this capacity, the situation cannot be rationally navigated in the first place.

Dreyfus neglected the human capacity to articulate new criteria for decisive actions. This allows people to not only navigate challenging situations, but to redraw their own moral horizons by challenging what is fundamentally at stake in a given activity. Such situations may require deep strategic thinking or building consensus before coming to a decision, or may simply be unprecedented. The point is that ethical intervention requires the capacity to reason–to apply symbols to the world in a way that defines available actions as good or bad. Being rational is critical to ethical decision-making. To be ethical is to take responsibility for one's actions, not merely following rules but intentionally applying them as a guide for behavior.

No matter how capable the system is, today's AI is prone to fail whenever there is more than one interpretation of the task–more than one set of possible rules–at hand. In these situations, humans must intentionally intervene and introduce new criteria to explicitly distinguish types of values or input labels. This is how to build into the system a sense of what is ethical and good.

### 4. OBSERVATION OR JUDGMENT IS ALWAYS MADE FROM THE POINT OF VIEW OF THE SUBJECT.

The popular conception of data is that it is objective and independent of theory or definitions of terms. But this is false in both the natural and social sciences. As elaborated by Kuhn in *The Structure of Scientific Revolutions*, the same piece of physical data has different meanings when viewed through the lens of different theories. Feyerabend shed further light on the nature of a data point. In *Against Method*, he wrote that the "[historico-physiological character of evidence] does not merely describe some objective state of affairs but also expresses subjective, mythical, and long-forgotten views concerning this state of affairs".

Any collection of data points is meaningless without a theory arranging it in a particular way. Even the decision to collect data requires some preliminary understanding of the world, some theory of how the world works that could be refuted or confirmed with more data. Hence, the very decision to collect additional data is theory-laden. As expressed by Ludwig von Mises: "Economic history is possible only because there is an economic theory capable of throwing light upon economic actions. If there were no economic theory, reports concerning economic facts would be nothing more than a collection of unconnected data open to any arbitrary interpretation."[36]

---

[35] Dreyfus, Hubert L. *What computers still can't do: A critique of artificial reason*. MIT press, 1992.

[36] Von Mises, Ludwig. *Human action, The Scholar's Edition*. Mises Institute, 2010.

The researcher must thus rely on interpretation based on theory, not impartiality, when collecting data: "The historian does not simply let the events speak for themselves. He arranges them from the aspect of the ideas underlying the formation of the general notions he uses in their presentation. He does not report facts as they happened, but only relevant facts. He does not approach the documents without presuppositions, but equipped with the whole apparatus of his age's scientific knowledge."[37]

But if all researchers agreed on an interpretation of data points, is objectivity possible? According to Friedrich Nietzsche's notion of perspectivism, the answer is no. Nietzsche claimed that moral philosophy always reflects the psychological needs of the philosopher espousing it: "Gradually it has become clear to me what every great philosophy so far has been: namely, the personal confession of its author and a kind of involuntary and unconscious memoir, also that the moral (or immoral) intentions in every philosophy constituted the real germ of life from which the whole plant had grown."[38]

For Nietzsche, this means that the true value of reasoning lies in its use as an instrument in moral struggles (the "will to power"), not in claims to objectivity. Repressing one's subjectivity implicitly devalues reason itself: Moral reasoning is not necessarily objective, it can be analyzed as an instrument of philosophers' will to power. To be clear, Nietzsche doesn't consider it in a negative light. For him, it's the only way to reason: "But to eliminate the will altogether, to suspend each and every affect, supposing we were capable of this–what would that mean but to castrate the intellect?"[39]

In other words, even if an objective world exists, Nietzsche suggests there is no such thing as an objective ethics. Our perception of reality is colored by our subjective interpretations of events, which in turn are colored by our values. People in turn will disagree about what is ethically good or reprehensible based on their own desires, fears, and psychic conflicts. Skill at reasoning allows one to move from angry incoherent outbursts and expressions of disgust to treatises on morality, but not to impartial truths. This conclusion is not cynical. To avoid nihilism, the philosopher must embrace his or her own ethical standpoint rather than hide behind false claims to objectivity.

---

[37] Von Mises, Ludwig. *Human action, The Scholar's Edition*. Mises Institute, 2010.
[38] Nietzsche, Friedrich. *Beyond Good & Evil: Prelude to a Philosophy of the Future*. Vintage, 1987. BGE 6.
[39] Nietzsche, Friedrich. *On the Genealogy of Morals and Ecce Homo*. Vintage, 1989. GM III,12.

# IMPLICATIONS: THE CURRENT DAIOS PRODUCT

### WHAT DOES OUR PRODUCT CURRENTLY DO?

daios is an AI ethics solution that teaches machines morality by co-creating algorithms with end users. The solution connects technical aspects that determine reality for machines, e.g. data, with the human element, such as AI development teams and those otherwise without a voice, the end-users.

The solution is built on top of a platform that uses AI, explaining how the current training data influences AI behavior, revealing any hidden ethical values, and giving users a place to give anonymized and secure feedback on algorithm behavior.

In ancient Greece and Rome, aqueducts were used to channel sources of water directly to cities, enabling them to grow and thrive.

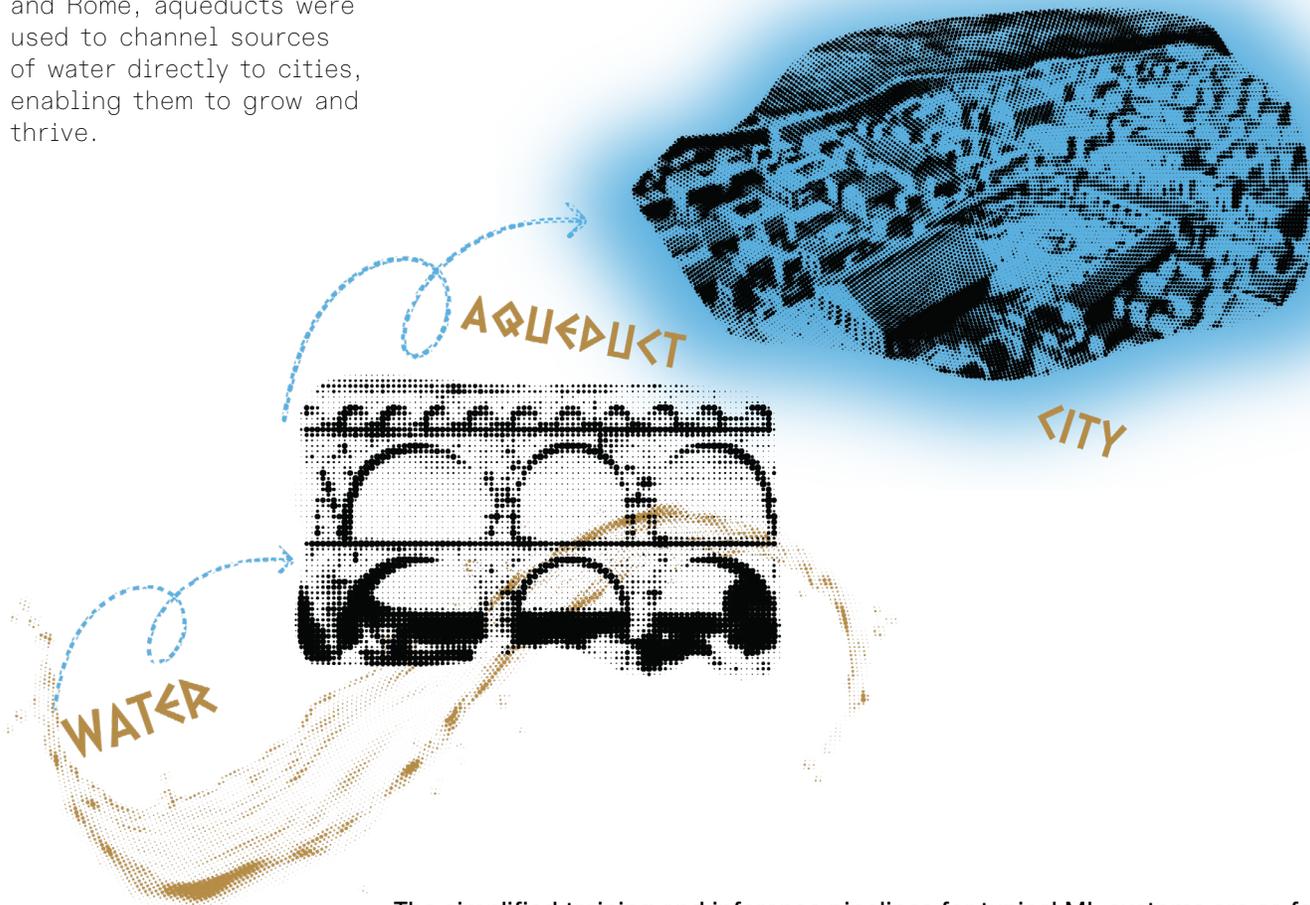

The simplified training and inference pipelines for typical ML systems are as follows.

### TRAINING STAGE:
1. Collect and label training data
2. Train
3. Use the output of the model to continue training until certain technical criteria are met

### DEPLOYMENT STAGE:
1. Input real data into the model
2. Run inference using the already trained model
3. Use the output for the AI-powered application

Most other AI ethics solutions tightly couple the ethics of AI and the training and deployment pipelines. This tight coupling unnecessarily conflates the theoretical and engineering challenges of training and deploying an AI system with ethics. daios is distinct from the usual training and deployment process. The ethical layer doesn't interfere with ML researchers' and engineers' usual work.

There are multiple approaches to implementing an ethical layer: supervised fine-tuning, reinforcement learning methods, and others. One state-of-the-art approach to AI ethics is the way taken by the OpenAI Alignment team. The team fine-tuned GPT-3 to align the model with human intent. The training set for fine-tuning was collected by human labelers that demonstrated the desired model behavior. OpenAI's attempt to be representative consisted of hiring labelers sampled from various genders, ethnicities, nationalities, ages, and educational levels. But the team only used the collected data that was already aligned with the values of the OpenAI research team itself. In fact, one of the hiring criteria for labelers was "agreement on sensitive speech flagging".[40] Rather than clarifying the ethical values built into GPT-3, the team focused on the diversity of perspectives within collected data, which does not solve the underlying problem. Nor will using other approaches like "Constitutional AI", which align the ethics of an AI system with the ethics of the team creating it.[41]

The "daios version" of AI alignment is for AI to be in tune with the values of users of AI systems, as opposed to in alignment with those who build or manage the system. The diagram below illustrates the fact that the source of ethical values is the user, not daios. This is the distinguishing feature of the platform. All other mentioned approaches aim at being "neutral", "harmless", or supporting variously defined "human values".

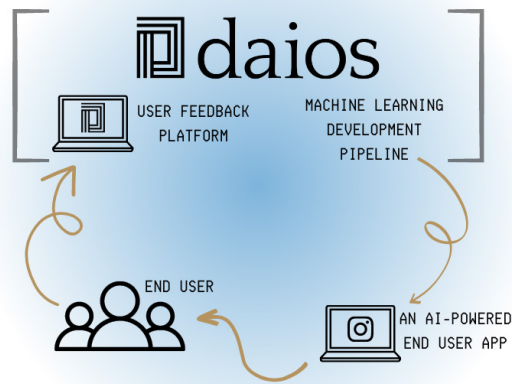

daios reuses some of the techniques developed by other players in the industry, but the platform remains grounded in the needs and wishes of our users. We will never develop AI systems in ways that conform merely to our own ethical views. Instead, we fine-tune algorithms with user feedback based on curated datasets and a unique IP. The resulting algorithms are both good for users and the companies that create them.

The world is permeated by diverse and complex ethical systems and should be reflected in the way users interact with daios. Rather than dictating values and building them into AI systems, daios offers an opportunity for users to participate in every decision an AI makes.

---

[40] Ouyang, Long, et al. "Training language models to follow instructions with human feedback." *arXiv preprint arXiv:2203.02155* (2022).

[41] Bai, Yuntao, et al. "Constitutional AI: Harmlessness from AI Feedback." *arXiv preprint arXiv:2212.08073* (2022).

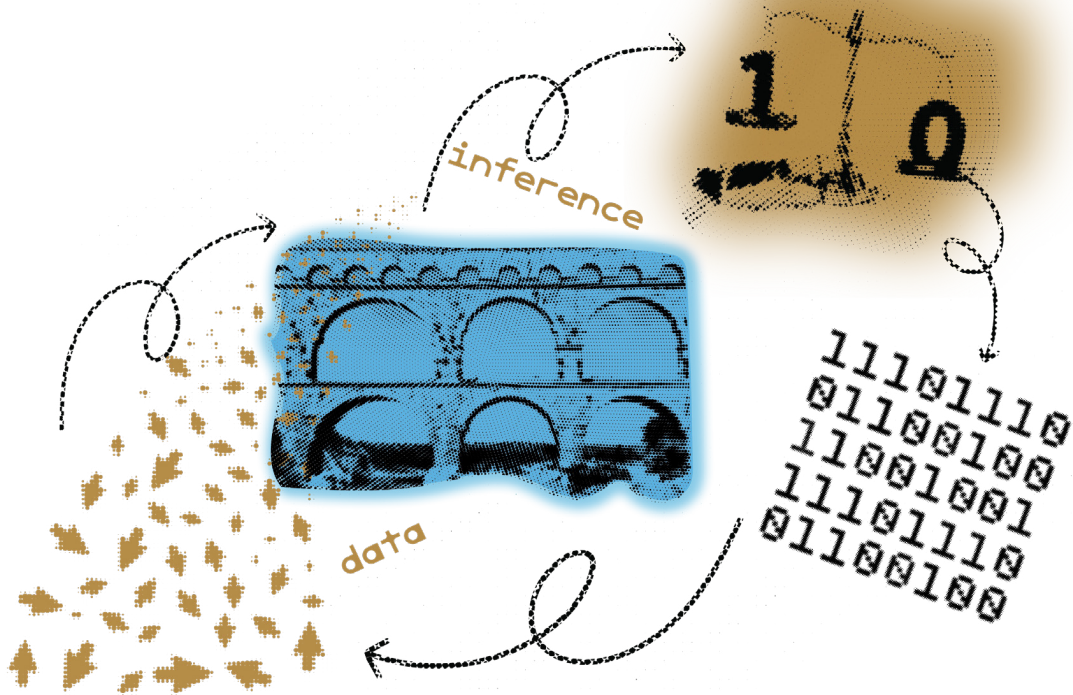

The aim of daios is to do for AI systems what aqueducts did for ancient cities: channel, filter, and distribute outputs for public benefit as part of a cohesive feedback loop.

In practice, daios is creating a user interface that emphasizes knowledge from the side of the user, i.e. direct moral, value-specific feedback to the AI system. The person interacting with and influenced by the outputs of the AI system will be able to make a more informed decision, and give direct input on how that AI is trained. The type of information revealed may include any personal bias from the perspective of the expert, i.e. more or less aggressive treatments prescribed by a medical doctor[42], as well as ethical bias, i.e. pro-communist, anti-communist; pro- or anti-feminist; etc. This also means that daios is uniquely value-agnostic.

But why can't AI be value-neutral to begin with? As discussed in the axioms for ethical AI, data cannot be objective. The world is always perceived from the perspective of a subject; it's impossible to create data independent from a particular viewpoint. Therefore, technology will always be laden with hidden ethical assumptions; in the case of machine learning and AI, these assumptions are amplified and magnified.[43]

Traditionally, such as in the case of expert systems, adding more data was better. A more complex and sophisticated ontology created a better representation of reality for the machine and therefore produced more accurate outputs. In the case of AI, adding more terabytes of training data or more parameters will also help the ontology of a system to be more reflective of reality, but will never solve the inherent problem of data being subjective.

Rather than attempting to "solve" or scale its way out of ethical problems, daios is making ethical AI more tractable. daios assumes that AI will always be value-laden and therefore should make those values visible to end users, rather than implicitly encode the values of the designers or other favored stakeholders into the system.

[42] Let's imagine an application using AI/ML to help diagnose patients with lung cancer based on MRI scans of the lungs of a patient. The algorithm was trained on images annotated by a group of data labelers, who are called subject matter experts. In this case, the labelers must either be radiologists themselves or following the instructions of a doctor (in the case with multiple doctors, labeling must converge on one opinion). Let's imagine that the doctors labels tend towards diagnosing lung cancer early, and typically request more biopsies of nodules found within lungs than other doctors. The daios platform would detect that type of bias within datasets used to train the algorithm and automatically reveal the "personal bias" to the end user, the patients. This helps the patient to decide if they wish to accept the opinions of the AI, or not.

[43] Mansoury, Masoud, et al. "Feedback loop and bias amplification in recommender systems." Proceedings of the 29th ACM international conference on information & knowledge management. 2020.

# CONCLUSION

We have argued that neither legal compliance nor scalable computation are viable for giving AI systems moral virtue. Rather, the path to ethical AI starts with data. Ethics must be built directly into the data on which AI models are trained. Only in this way will the system reflect the subjectivities of those most impacted by its performance and whom it is meant to serve.

The tension between ethical commitments and technical capabilities–between what can and should be done–is inevitable. This is hardly a new problem, as any transformative technology from nuclear fission to social media raises new questions about how it should and should not be used. What is new with AI is the capacity to redesign human activities from scratch–how we get to work, order groceries, find life partners, and communicate.

While frightening, this tension can be resolved by putting our own values into the data on which AI is trained. As we observe the AI's subsequent performance, we remake those activities and learn more about what it is we really want. A positive feedback loop can then emerge between our assumptions about what is good and how the system learns from us over time. Putting ethics into data is therefore not about making AI conform to a rigid moral scheme–it's about becoming better, more fully-realized versions of ourselves.

# APPENDIX 1: GLOSSARY OF TERMS

1. ETHICS — a branch of philosophy that determines right and wrong by analyzing and defending different concepts of values.

2. USER FEEDBACK — qualitative and quantitative information from customers on their likes, dislikes, impressions, and preferences on a product or service. May take the form of email, surveys, third party research, in-app messaging, observation, etc.

3. TRAINING DATA — the data used to train machine learning models to predict the outcome you design your model to predict.

4. DATA LABELING — also, data annotation. The activity of manually assigning context or meaning to data to guide machine learning algorithms to achieve a desired output. Data labels are considered "ground truth" for algorithms.

5. MACHINE LEARNING PIPELINE — a series of steps taken to develop, train, deploy, and monitor a machine learning model.

6. END USERS — any user of a machine learning system.

7. MLOPS — machine learning operations, a core function of machine learning focused on streamlining the process of taking machine learning models into production (deployment), then maintaining and monitoring them.

8. KPI — key performance indicator. A quantifiable measure of performance over time for a specific objective. KPIs are used by businesses, especially enterprise companies, to give teams goals, measure progress, and indicate success or failure.

# APPENDIX 2: AI DEVELOPMENT TEAM INTERVIEWS

In order to better understand how engineering teams build AI/ML systems–and point out what is missing from existing practices–we conducted user interviews. Interviews were 30 minutes or less in length unless otherwise noted. Criteria for these interviews, and justification for those criteria, are listed below.

### REQUIREMENTS

- Teams working on AI/ML systems that have already been developed and standardized.

    Rationale:
    We wanted to speak with teams that already were having issues with AI behavior, after the system had matured to some extent. A system must exist first for us to make changes to it. Change in AI behavior is our key metric for determining performance.

    We cannot systematically measure change in AI behavior if the system is not repeatable, since we cannot be certain that the changes made were caused by our interference or by another cause.

- Machine learning systems that have been deployed.

    Rationale:
    While there are AI systems that use hard-coded rules, we did not pursue companies developing them. This is because for these systems, the solution to misbehavior is to add an additional a priori rule rather than improve over time through data-driven induction. Although many of these systems intermingle, we want to focus on one type of AI for the time being: machine learning and deep learning.

    Also, some of the ML engineers we spoke with were working on models that have not been deployed yet, so did not experience the data-driven problems that concern us. Pre-deployment models are often created using a public dataset and almost always need adjustment or retraining after being deployed.

- Machine learning systems that interact with human end users.

    Rationale:
    Some ML systems only interact with other automated components. By focusing on ML systems that interact with human end users, we are able to access and better under stand the critical situations in which human ethics and machine behavior interact.

### IDENTIFIED PAIN POINTS
We asked respondents what made changing AI behavior difficult.

#### 1. AI ETHICS FRAMEWORKS ARE TOO BULKY.
One AI/ML Product Manager from H&M's branch based in Sweden mentioned that whenever the recommender system did something unethical, the PM would call the team together and try to determine the best course of action.

The company had an in-house AI ethics team member, who was not familiar with the inner workings of the team or the system itself. Furthermore, AI ethics took the form of a checklist, which the Product Manager would need to check during product development rather than apply in critical situations.

### 2. INABILITY TO MEASURE THE IMPACT OF ETHICS METHODS.

A Responsible AI team member from H&M articulated the lack of specific KPIs attached to AI ethics: "Our biggest struggle right now is figuring out meaningful KPIs. At the moment, we do not have a way to directly measure the change of output in machine learning algorithms." The core issue was how to evaluate the methods she was using with evidence–just philosophical assumptions–that empirically demonstrated their effectiveness.

Because AI systems tend to be much faster than humans at decision-making it becomes very inefficient to manually measure change in AI outputs. This means the solution must be technical, if not also AI-based. In fact, there are many machine learning models that can monitor the output of another machine learning model. Alternatively, a rules-based model may be built, but again one runs into the issue of scalability. Over the long term, using machine learning to monitor machine learning systems will be critical to monitor the behavior of the system.

### 3. DETERMINING THE VALUES THAT SHOULD BE INCORPORATED INTO AI.

Another Responsible AI lead member mentioned that their team was using a framework for governance but struggled with building a team consensus on the kinds of values that should be incorporated into AI. Everyone had multiple values. It was challenging to know what to do.

This issue also appears from the user's standpoint. Rumman Chowdhury, the former Director of Machine Learning Ethics, Transparency, and Accountability (META) at Twitter was interviewed on opportunities and challenges in the field of AI ethics: "The single biggest advancement I'm looking forward to is improvement in algorithmic choice for users. That means offering individuals increased agency, control, understanding, and transparency over the computational systems that govern their experience."[44] Implementing this concept in practice presents a true challenge, as the concept of "choice" refers not merely to a binary preference, such as "a" or "b", but also to the determination of value itself.

### 4. MODELS UNABLE TO FUNCTION IN MULTIPLE CONTEXTS.

From a technical perspective, multiple AI engineers cited the lack of appropriate data to retrain or adapt the model to work in a new context. One Product Manager revealed that millions were spent at Yelp to solve this issue, while an AI engineer developing a voice system for McDonald's mentioned this problem as well.

### PRODUCT – NEEDED FEATURES

What features do we need to have in order to build a product that fits the demands of the current market?

First, we need a solution that scales with AI system development. When we use the word scale, we are referring to the ability of companies to increase revenue faster than costs. Machine learning as a technology is scalable but there is a hidden bottleneck in data labeling, which is labor-intensive, expensive, and time consuming. Current governance frameworks for ethical AI development are too bulky. Frameworks usually end up taking the backseat to product development and when problems do arise, engineers tend to sidestep frameworks in favor of a faster solution. The core issue is that the use of frameworks becomes updating checklists, rather than a dynamic and interactive solution in the same way AI is.

Second, the product needs to give access to metrics (i.e. KPIs) that indicate value. There is a simple way to solve this problem, which is to start measuring the ethical output of ML systems. Many MLOps startups already are tackling this issue.

---

[44] https://www.emergingtechbrew.com/stories/2022/01/17/seven-ai-ethics-experts-predict-2022-s-opportunities-and-challenges-for-the-field

Third, ethics should be determined by those most affected by the AI system, which are major stakeholders (i.e. the creators, the developers, designers, etc) but, most of all, the end users. At this moment in time, the scales are tipped in favor of the machine developers and companies, rather than the people that actually use the technology. This is one of our most essential insights: creating a way to systematically incorporate meaningful human interaction into the machine behavior itself. Automated systems are designed to make decisions at a speed that greatly outpaces human decision-making, which means we must find another way to meaningfully integrate human interaction with machine learning.

### FINAL THOUGHTS

The four themes of our interviews outlined above reveal a common thread: the need to weave ethical commitments directly into the data flow. Only in this way can the behavior of the system itself be made observable, able to be evaluated, and changeable with respect to ethical concerns. The problem is that this technical capability does not exist in present machine learning pipelines. This makes ethical problems a source of ongoing frustration rather than an opportunity to improve behavior over time.

# ACKNOWLEDGEMENTS

For their helpful comments and suggestions on this paper, the authors wish to thank Jon Anderson, John Basl, Moritz Bierling, Micah Carroll, Victoria Celano, Kathleen Creel, Tyna Eloundou, Jeremy Galen, Abhishek Gupta, Alex Jozsa, Seth Lazar, Solveig Neseth, Tung Nguyen, Terence Park, Ella Peters, Travers Rhodes, Cat Struss, Daniel Susser, Nataleigh Waters, Magda Welle, Lily Xu, Qian Yang, and J.D. Zamfirescu-Pereira. The authors also thank the Tech Ethics Lab at the University of Notre Dame for supporting the research resulting in and drafting of this paper.

Additionally, we want to thank all the participants of our user interviews for your invaluable and continual insights. We thank Loes Claessens for designing the infographics and page layout.

# ABOUT THE AUTHORS

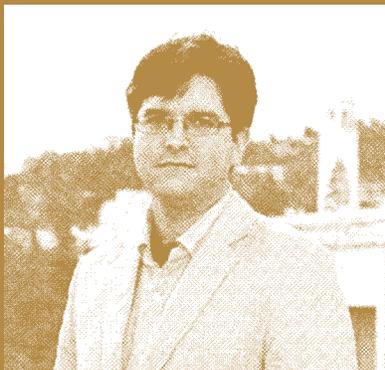

Thomas Krendl Gilbert is a postdoctoral fellow at Cornell Tech in New York City, and AI Ethics Lead at daios. He previously designed and received a Ph.D. in Machine Ethics and Epistemology at the University of California, Berkeley. Thomas researches the emerging political economy of AI, in particular reinforcement learning systems.

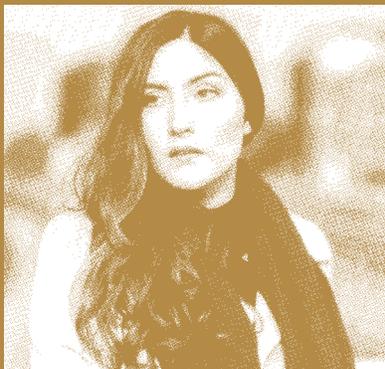

Megan is the CEO and co-founder of daios. She has been working in AI/ML startups for over 6 years, specializing in operations, content marketing, and strategy. She has been COO twice. Megan has an MA in Philosophy focused on the methodology of science, which inspired the daios solution.

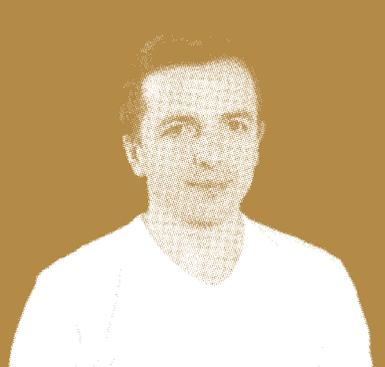

Andrew is the CTO and Head of Product at daios. He has been building AI/ML systems since 2012. He established and scaled data labeling operations in two organizations.

daios